\documentclass{article}
\usepackage{spconf}

\usepackage{appendix}
\usepackage{makecell}
\usepackage{enumitem}

\usepackage[english]{babel}

\usepackage{ifpdf}

\usepackage{cite} 
\usepackage{url}
\usepackage{hyperref}
\usepackage{enumitem}

\usepackage[pdftex]{graphicx}
\graphicspath{{./figures/}}
\usepackage{color}
\usepackage{pgf, tikz, pgfplots}
\usetikzlibrary{shapes, arrows, automata, plotmarks}
\usetikzlibrary{calc,hobby,decorations}
\usepackage{fixltx2e}
\usepackage{stfloats}
\usepackage{tikz}
\usetikzlibrary{positioning}

\usepackage[cmex10]{amsmath}
\usepackage{amsfonts, amssymb, amsthm}
\usepackage{mathrsfs}

\usepackage{algorithm,algpseudocode}
\usepackage{booktabs}       
\algnewcommand{\LeftComment}[1]{\Statex \(\triangleright\) #1}

\usepackage{color, colortbl}
\definecolor{Gray}{gray}{0.9}
\usepackage{array}

\usepackage{enumitem}

\usepackage{multirow}
\usepackage{subcaption}

\input{pennColors.sty}
\usepackage{needspace}

\input{mySymbol.sty}

\newcommand{\QED}{\hfill\ensuremath{\blacksquare}}

\title{HODGE-AWARE CONTRASTIVE LEARNING}
\name{Alexander Möllers$^{*}$, Alexander Immer$^\dagger$ , Vincent Fortuin$^\ddagger$, Elvin Isufi$^{*}$ \vspace{-1.5cm} \thanks{${*}$ TU Delft, Delft, The Netherlands. $\dagger$ ETH Zürich, Zürich, Switzerland. $\ddagger$ Helmholtz AI, Munich, Germany. EI is supported by the TU Delft AI Labs Programme. AI is supported by a Max Planck ETH Center for Learning Systems doctoral fellowship. VF was supported by a Branco Weiss Fellowship. AM conducted the work while at ETH Zürich. Correspondence to alexander.j.moellers@gmail.com. This work has been submitted to the IEEE for possible publication. Copyright may be transferred without notice, after which this version may no longer be accessible. }
\address{~}}

\begin{document}
\ninept
\maketitle

\begin{abstract}

Simplicial complexes prove effective in modeling data with multiway dependencies, such as data defined along the edges of networks or within other higher-order structures. Their spectrum can be decomposed into three interpretable subspaces via the Hodge decomposition, resulting foundational in numerous applications. We leverage this decomposition to develop a contrastive self-supervised learning approach for processing simplicial data and generating embeddings that encapsulate specific spectral information.
Specifically, we encode the pertinent data invariances through simplicial neural networks and devise augmentations that yield positive contrastive examples with suitable spectral properties for downstream tasks. Additionally, we reweight the significance of negative examples in the contrastive loss, considering the similarity of their Hodge components to the anchor. By encouraging a stronger separation among less similar instances, we obtain an embedding space that reflects the spectral properties of the data.
The numerical results on two standard edge flow classification tasks show a superior performance even when compared to supervised learning techniques. Our findings underscore the importance of adopting a spectral perspective for contrastive learning with higher-order data.

\end{abstract}
\begin{keywords}
Hodge Laplacian, simplicial filter, contrastive learning. \vspace{-.5cm}\vskip-.5cm
\end{keywords}


\section{Introduction}\vskip-.2cm
\label{sec:intro}

The difficulty to represent many biological, social, and technological networks arises from the complexity of their inherent relational structures and the nuanced definitions of the associated data. Importantly, the data in such networks is often defined on higher-order components like edges and triangles, thus demanding an approach that incorporates interactions beyond the pairwise paradigm \cite{BATTISTON_20201_pairwise}. Simplicial complexes (SCs) have been proposed to model such dependencies in higher-order networks \cite{robinson2014topological,schaub2021signal}. Amongst others, SCs have shown particular efficacy in edge flow applications (e.g., mass, energy, information, or trajectories) for which traditional graph-based techniques do not induce a good inductive bias \cite{barbarossa2020topological,schaub_hodgelets2022}. Notably, they have been leveraged to alleviate the curse of dimensionality in autoregressive flow prediction for water networks \cite{isufi_2023_svam} and to remove arbitrage opportunities in currency exchange markets \cite{yang_2022_sc_filters}.

An important property of SCs is that they enjoy an algebraic representation via the Hodge Laplacian matrices, ultimately allowing for spectral analysis. The latter is achieved through the Hodge decomposition of the spectrum and has been used to develop Hodge-aware signal processing and learning techniques for edge flows and other simplicial data \cite{Yang_2021_finite,yang_trendfilter_2022,schaub_hodgelets2022}. Recent advances include developing a simplicial Fourier transform \cite{barbarossa2020topological}, convolutional filters \cite{yang_2022_sc_filters}, and neural networks that are trained in a (semi-)supervised manner. \cite{ebli2020simplicial,roddenberry2021principled,bodnar2021weisfeiler,simp_con_networks_elvin}.

While these supervised learning (SL) methods for simplicial data have their merits, they also come with limitations such as a over-reliance on labeled examples and suboptimal performance in data-scarce scenarios. SL methods are also mainly designed for a specific task and the resulting output is generally not reusable in different applications. Therefore, we propose a contrastive learning (CL) approach for simplicial data that addresses these issues. To this end, we employ a simplicial convolutional neural network (SCNN) to produce embeddings and optimise it in a self-supervised manner using the InfoNCE loss \cite{Oord2018}.  The resulting CL approach can use both labeled and unlabeled examples to create robust and reusable representations for simplicial data. These representations can subsequently be employed in various downstream tasks, offering improved accuracy especially in label-scarce scenarios.

To further enhance our method, we propose stochastic augmentations and introduce information about the Hodge Decomposition into the embeddings by i) optimizing the parameters of the augmentations so that they generate positive examples that respect the spectral properties of the data; and ii) weighting the negative samples in the InfoNCE loss by a Hodge-aware distance to the anchor (true datum). This approach results in a spectrally organized embedding space and facilitates downstream learning. We corroborate the latter in two edge flow classification settings and show superior performance compared with a fully-supervised model. Related to this, our contribution is threefold:

\begin{enumerate}[label=\textbf{C\arabic*)}, start = 1]
    \item we propose simplicial contrastive learning (SCL), design related augmentations and experimentally validate all our approaches;
    \item we show how augmentations in the simplicial domain can be optimized with respect to the Hodge decomposition;
    \item we introduce a reweighing of the negative examples based on the similarity of their Hodge components to encourage a spectrally organized embedding space.  
\end{enumerate}
%


\section{Data Processing on Simplicial Complexes}\vskip-.2cm
\label{sec:prelim}

In this section, we introduce the fundamental concepts behind simplicial data structures and related data processing techniques. \vspace{-0.2cm}

\subsection{Simplicial Data and Structures}\vspace{-.1cm}

Let $\ccalV = \{1, \ldots, N\}$ be a set of vertices. A $k-$simplex $\ccalS^k$ is a subset of $\ccalV$ that contains $k+1$ distinct elements. A simplicial complex $\ccalX^K$ of order $K$ is a collection of simplices such that it contains at least one $K-$ simplex and if $\ccalS^k \in \ccalX^K$ we have that all subsets of $\ccalS^k$ are also elements of $\ccalX^K$. From a geometric representation perspective, in a SC of order $K=2$, nodes are $0$-simplices, edges are 1-simplices and (filled) triangles are 2-simplices \cite{robinson2014topological,schaub2021signal}.


Neighborhood relations in an SC can be represented via the incidence matrices and Hodge Laplacians. Specifically, the incidence matrix $\mathbf{B}_k \in \mathbb{R}^{N_{k-1} \times N_k}$ captures the adjacencies between $(k-1)-$ and $k-$simplices. Consequently, $\mathbf{B}_1$ represents the node-to-edge incidence matrix and $\mathbf{B}_2$ the edge-to-triangle incidence matrix. For an SC of order two, $\ccalX^2$, the related Hodge Laplacians are defined as:

\begin{equation*}\begin{array}{l} {{\mathbf{L}}_0} = {{\mathbf{B}}_1}{\mathbf{B}}_1^ \top , \\ {{\mathbf{L}}_1} = {{\mathbf{L}}_{1,\ell }} + {{\mathbf{L}}_{1,u}}: = {\mathbf{B}}_1^ \top {{\mathbf{B}}_1} + {{\mathbf{B}}_2}{\mathbf{B}}_2^ \top , \\ {{\mathbf{L}}_2} = {\mathbf{B}}_2^ \top {{\mathbf{B}}_2} \end{array}\tag{2}\end{equation*}
where $\mathbf{L}_0$, $\mathbf{L}_1$, $\mathbf{L}_2$ represent the neighborhood relationships between nodes, edges, and triangles, respectively. Moreover, $\mathbf{L}_0$ coincides with the standard graph Laplacian \cite{schaub2021signal}. The lower-Laplacian $\mathbf{L}_{1, \ell}=\mathbf{B}_1^{\top} \mathbf{B}_1$ and upper-Laplacian $\mathbf{L}_{1, u}=\mathbf{B}_2 \mathbf{B}_2^{\top}$ split the edge adjacencies into relations that are due to common vertices and common triangles, respectively.


A $k-$simplicial signal $x^k$ is a mapping from a $k-$simplex to the set of real numbers that formalizes the simplicial data. We collect all $k-$ signals in vector $\bbx^k = [x^k_1, \ldots, x^k_{N_k}]^\top$, where $x_i^k$ is the signal on the $i$th simplex and $N_k$ is the total number of $k-$simplices. For instance, we denote an edge flow as $\bbx^1 = [x^1_1, \ldots, x^1_{N_1}]^\top$ with $x^1_e$ being the flow on the edge $e=(m, n)$ in $\mathcal{S}^1$. In the sequel, we focus on edge flows to ease exposition and because of their wider applicability; thus we drop the superscript and denote them as $\bbx$.

\vspace{-.15cm}
\subsection{The Hodge Decomposition}\vspace{-.1cm}

SCs allow for a spectral processing of simplicial signals via the Hodge decomposition, which decomposes the space $\reals^{N_k}$ as: 

\begin{equation}\label{eq.spaceDecomp}
\reals^{N_k} = \text{im}(\bbB_k^\top) \oplus \text{im}(\bbB_{k+1}) \oplus \text{ker}(\bbL_k) 
\end{equation}
where $\text{im}(\cdot)$ and $\text{ker}(\cdot)$ are the image and kernels spaces of a matrix and $\oplus$ is the direct sum of vector spaces \cite{barbarossa2020topological,yang_2022_sc_filters}. Accordingly, we can decompose any edge flow $\bbx$, into three parts $\bbx = \bbx_{\rm G} + \bbx_{\rm C} + \bbx_{\rm H}$ each living in an orthogonal subspace known as the gradient space $\bbx_{\rm G} \in \text{im}(\bbB_1^\top)$, the curl space $\bbx_{\rm C} \in \text{im}(\bbB_{2})$, and the harmonic space $\bbx_{\rm H} \in \text{ker}(\bbL_1)$. In turn, the 1-Hodge Laplacian can be eigendecomposed as $\bbL_1 = \bbU \bbLambda \bbU^\top$ with eigenvectors $\bbU$ and eigenvalues $\bbLambda$. By grouping the eigenvectors as $\bbU = [\bbU_{\rm G}, \bbU_{\rm C}, \bbU_{\rm H}]$ and projecting the edge flow onto them gives rise to the embeddings $\tilde{\mathbf{x}}_{\mathrm{G}}=\mathbf{U}_{\mathrm{G}}^{\top} \mathbf{x}$, $\tilde{\mathbf{x}}_{\mathrm{C}}=\mathbf{U}_{\mathrm{C}}^{\top} \mathbf{x}$, $\tilde{\mathbf{x}}_{\mathrm{H}}=\mathbf{U}_{\mathrm{H}}^{\top} \mathbf{x}$. These embeddings encode different interpretable properties of the data, which we will exploit in Sec.~\ref{sec:methods} to design augmentations. Refer to \cite{barbarossa2020topological} for more detail on the role of the Hodge decomposition on processing simplicial data.

\vspace{-.15cm}
\subsection{Simplicial Convolutional Neural Networks}\vspace{-.1cm}\label{sec:scnn}

Simplicial convolutional filters are linear parametric mappings that can process simplicial signals \cite{yang_2022_sc_filters}. For an input edge flow $\bbx$, the filtered signal is $\bby =  \bbH(\bbL_1)\bbx$ with simplicial filtering matrix:
\begin{equation}\label{eq.simplFilter}
 \bbH(\bbL_1):= \bigg(\epsilon \mathbf{I} + \sum_{l_1 = 0}^{L_1}\alpha_{l_1}\bbL_{1,\ell}^{l_1} + \sum_{l_2 = 0}^{L_2}\beta_{l_2}\bbL_{1,u}^{l_2}			\bigg).
\end{equation}
Here, $\{\epsilon, \alpha_{l_1}, \beta_{l_2}\}$ are filter coefficients and $\mathbf{L}_{1, \ell}$, $\mathbf{L}_{1, u}$ propagate the edge signal via their respective neighbourhood relations. The filter is local in the sense that it moves information by at most $L = \max\{L_1, L_2\}$ hops across the simplicial structure. Moreover, by exploiting the recursions $\bbL_{1,\ell}^{l_1}\bbx = \bbL_{1,\ell}(\bbL_{1,\ell}^{l_1-1}\bbx)$,  $\bbL_{1,u}^{l_2}\bbx^k = \bbL_{ku}(\bbL_{1,u}^{l_2-1}\bbx)$ and since the Laplacian matrices are sparse, we can obtain the output with a cost of order $\ccalO(LN_1)$. The filter part related to the lower Laplacian acts on the signal gradient embedding, whereas that related to the upper Laplacian acts on the curl embedding. All the parameters contribute to processing the signal harmonic embedding. Refer to \cite[Sec. IV]{yang_2022_sc_filters} for the specific relation of filter \eqref{eq.simplFilter} and decomposition \eqref{eq.spaceDecomp}.

The constant number of parameters and the linear computational complexity in the number of edges, make the filter \eqref{eq.simplFilter} an appealing solution to learn representations from edge flows. To also learn non-linear mappings, simplicial convolutional neural networks (SCNNs) have been developed as layered structures interleaving filters with pointwise non-linearities \cite{simp_con_networks_elvin}. Specifically, given the SCNN input $\bbx_0 :=\bbx$, the propagation rule at each layer $t$ is:
\begin{equation} \label{eq:encoder}
\mathbf{x}_{t} = \sigma\big(\bbH_t(\bbL_1)\bbx_{t-1} \big)
\end{equation}
where $\sigma(\cdot)$ is a pointwise nonlinearity (e.g., ReLU). The final layer constitutes the SCNN output which provides its embedding. Compactly, we will denote the SCNN input-output relation as $\bbh :=f_\ccalH(\bbx)$, where $\bbh$ is referred to as the SCNN embedding and set $\ccalH := \{\epsilon_t, \{\alpha_{\ell_1,t}\},\{\beta_{\ell_2,t}\}\}_t$ collects the parameters of all layers. Furthermore, depending on the setting, a readout function is applied to $\bbh$ to transform it for the task at hand (e.g., binary classification).

\vspace{-.2cm}
\section{Problem Statement}\vspace{-.15cm}

Learning representations via the SCNN is typically done in a supervised manner, but we often have just a few labeled examples or we are missing labels at all. To tackle this challenge, we resort to contrastive learning and propose a self-supervised learning approach for simplicial complexes. Our problem statement reads as:

\textit{Given a set of unlabeled edge flows, we want to train a simplicial convolutional neural network in a self-supervised manner to generate embeddings that reflect the Hodge-properties of the data and that can be used in a downstream task.}

We approach this problem by training the network with the contrastive InfoNCE loss and by designing augmentations that preserve the desired spectral properties. We also reweight the negative samples in the loss to push apart spectrally-different embeddings.


\begin{figure}[t!]
  \centering
  
\begin{tikzpicture}[
roundnode/.style={circle, draw=black!80, fill=white!10, very thick, minimum size=7mm},
squarednode/.style={rectangle, draw=black!80, fill=blue!5, very thick, minimum size=5mm, minimum width=1.2cm},
]
\node[roundnode]      (h1)                              {$\mathbf{h}_{1}$};
\node[roundnode]        (z1)       [above=0.5cm of h1] {$\mathbf{z}_{1}$};
\node[roundnode]        (x1)       [below=0.8cm of h1] {$\mathbf{x}_{1}$};
\node[roundnode]      (h2)       [right=2.5cm of h1] {$\mathbf{h}_{2}$};
\node[roundnode]        (x2)       [right=2.5cm of x1] {$\mathbf{x}_{2}$};
\node[roundnode]        (z2)       [right=2.5cm of z1] {$\mathbf{z}_{2}$};
\node[roundnode]        (x)        [below=of $(x1)!0.5!(x2)$] {$\bbx$};
\node[squarednode]      (infonce)        at ($(z1)!0.5!(z2)$) {InfoNCE};  

\draw[->] (h1.north) --  (z1.south) node[midway, left=1mm]{$g_\ccalH(\cdot)$};
\draw[->] (h2.north) --  (z2.south) node[midway, right=1mm] {$g_\ccalH(\cdot)$};

\draw[->] (x1.north) -- (h1.south) node[midway, left=1mm] {$f_\ccalH(\cdot)$};
\draw[->] (x2.north) -- (h2.south) node[midway, right=1mm] {$f_\ccalH(\cdot)$};
\draw[->] (x.north) -- (x1.south)  (x1.south) node[pos=0.2, left=5mm]{$\ccalT_{1}(\cdot)$
};
\draw[->] (x.north) -- (x2.south) (x2.south) node[pos=0.2, right=5mm]{$\ccalT_{2}(\cdot)$};
\draw[->] (infonce.west) -- (z1.east);
\draw[->] (infonce.east) -- (z2.west);
\end{tikzpicture}

\vskip-.15cm \caption{InfoNCE learning with augmentations. The data point (anchor) $\bbx$ submits two transformations $\ccalT_{1\backslash 2}(\cdot)$ to generate positive augmented examples. The latter are first passed through the SCNN $f_\ccalH(\cdot)$ to generate the simplicial embeddings $\bbh$ (cf. \eqref{eq:encoder}) and then to a parametric map $g_\ccalH(\cdot)$ to obtain the final representation $\bbz$. These representations are contrasted in loss \eqref{eq.infoNCEloss} to train both $f_\ccalH(\cdot)$ and $g_\ccalH(\cdot)$.}\vskip-.15cm
 
   \label{fig:sscl_struc}
\end{figure}
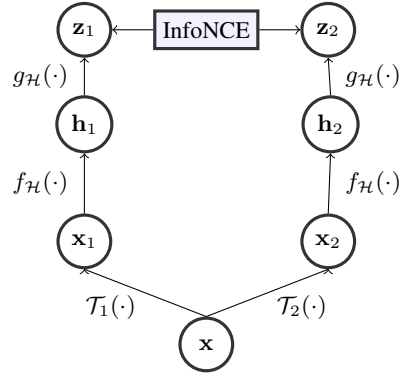


\vspace{-.2cm}
\section{Simplicial Contrastive Learning}\vspace{-.15cm}
\label{sec:methods}

To train an SCNN in a self-supervised manner, we resort to the contrastive learning framework \cite{SimCLR,graphcl_2020}. In the simplicial setting, this principle suggests that for each edge flow datum $\bbx$ we create both positive and negative examples and train the SCNN (a.k.a. the encoder) to map the positive embeddings close to each other and the negative ones farther apart. Specifically, we consider an edge flow datum $\bbx$ with its final representation $\bbz = g_{\ccalH}(\bbh) = g_\ccalH(f_\ccalH(\bbx))$, where $f_{\ccalH}(\cdot)$ is the SCNN [cf. \eqref{eq:encoder}] and $g_\ccalH(\cdot)$ is a parametric map (typically a fully connected layer). Then, we create a positive pair for $\bbx$ via two topological augmentations $\mathbf{x}_{i}^{\prime}= \mathcal{T}_{1}(\mathbf{x})$ and $\mathbf{x}_{j}^{\prime} = \mathcal{T}_{2}(\mathbf{x})$ with respective representations $\bbz_i^\prime = g_\ccalH(f_\ccalH(\bbx_i^\prime))$ and $\bbz_j^\prime = g_\ccalH(f_\ccalH(\bbx_j^\prime))$. The negative examples w.r.t. to $\bbx$ consists of other edge flows from the dataset $\bbx_m \neq \bbx$ and their augmentations. With reference to Fig.~\ref{fig:sscl_struc}, the overall network is trained to minimize the so-called temperature-scaled InfoNCE objective:
\begin{equation} \label{eq.infoNCEloss}
\mathcal{L}_{\text {infoNCE }}=-\sum_{\left(\mathbf{x}_{i}^{\prime}, \mathbf{x}_{j}^{\prime}\right) \in \mathbb{P}} \log \left(\frac{e^{{\rm sim}\left(\mathbf{z}_i^{\prime}, \mathbf{z}_j^{\prime}\right) / \tau }}{\sum_{m=1}^M e^{{\rm sim}\left(\mathbf{z}^{\prime}_i, \mathbf{z}_m\right) / \tau }}\right)
\end{equation}
where $\mathbb{P}$ is the set of all positive pairs in the data, $\operatorname{sim}(\boldsymbol{u}, \boldsymbol{v})=\boldsymbol{u}^{\top} \boldsymbol{v} /\|\boldsymbol{u}\|_2\|\boldsymbol{v}\|_2$ is the cosine similarity, $\tau$ is a temperature parameter, and  $M$ is the number of negative examples $\bbx_m$  with representations $\bbz_m$. The numerator encourages the network to map the positive embeddings close to each other, while the denominator repulses the negative embedding $\bbz_m$ from the positive one $\bbz_i^\prime$. 

The InfoNCE optimizes a lower bound on the mutual information between the representations of the positive pairs \cite{Oord2018}. Hence, augmentations should only preserve the information necessary to perform well on a downstream task \cite{isola_2020_infomin}. Consequently, the irrelevant information is destroyed and the mutual information encoded in the representations is optimized for the task at hand. Common augmentations in contrastive learning are stochastic and include masking part of the data (pixels, vertices, edges, node features, etc.), which expresses the belief that the most important parts of the data are preserved when a few connections or feature values are removed  \cite{graphcl_2020,grace_zhu_2020}. While this preserves, in probability, crucial information and works well empirically, note that the most relevant parts of a data point may not always remain intact. To mitigate this, the dropout probabilities are often chosen such that the most important properties of the data are conserved more often (e.g., in graphs nodes with a high centrality) instead of randomly removing information \cite{Zhu_2021,liu2022revisiting,lin2023spectral,graph_data_aug_surv_kaize_2022}. With this in place, we propose the following augmentation method.


\smallskip\noindent
\textbf{Edge flow masking.} This method masks edge flows with probabilities $\mathbf{p}$ to generate a positive example. That is,  $\mathbf{x'} = \ccalT(\bbx) := \mathbf{x} \circ \mathbf{e}$ where $\mathbf{e}$ is a random Bernoulli vector with entry $\mathbf{e}_{i} \sim Ber(\mathbf{p}_{i})$ and $\circ$ is the elementwise product. The standard approach is to pick the same masking probability for all edges. 

Such an augmentation is more effective in settings with binary flow types $\{-,1,1\}$ or when a zero value is not indicative. This is because the masked flow attains a zero value and the augmentation effectively drops parts of the flow. Next, we show how to optimize the masking probabilities w.r.t. flow Hodge-related information.

\vspace{-.1cm}
\subsection{Hodge-Aware Spectral Augmentations}

 Simplicial data often have particular properties in one of the three Hodge embeddings \cite{schaub2021signal,hdd_survey_2013}, which may be wrongly affected by augmentations if ignored. Hence, following the InfoMin principle simplicial augmentations should destroy information on irrelevant Hodge embeddings and preserve it on the others. To account for the latter, we cast the dropout probabilities as a stochastic optimization problem, considering the expected value of the difference of the generated Hodge embeddings to the embeddings of the anchor.  Specifically, consider the embeddings of an augmented example $\tilde{\mathbf{x}}^{\prime}_{\text{G}}=\mathbf{U}_{\mathrm{G}}^{\top}\mathbf{x}^{\prime}$,
 $\tilde{\mathbf{x}}^{\prime}_{\text{C}}=\mathbf{U}_{\mathrm{C}}^{\top}\mathbf{x}^{\prime}$,
 $\tilde{\mathbf{x}}^{\prime}_{\text{H}}=\mathbf{U}_{\mathrm{H}}^{\top}\mathbf{x}^{\prime}$. Then the expressions $\mathcal{L}_{\mathrm{G}}(\mathbf{p}) = \mathbb{E} [\left\|\tilde{\mathbf{x}}_{\mathrm{G}} -\tilde{\mathbf{x}}^{\prime}_{\text{G}} \right\|_2^2]$, $\mathcal{L}_{\mathrm{C}}(\mathbf{p}) = \mathbb{E} [\left\|\tilde{\mathbf{x}}_{\mathrm{C}} -\tilde{\mathbf{x}}^{\prime}_{\text{C}} \right\|_2^2 ]$, $\mathcal{L}_{\mathrm{H}}(\mathbf{p}) = \mathbb{E} [\left\|\tilde{\mathbf{x}}_{\mathrm{H}} -\tilde{\mathbf{x}}^{\prime}_{\text{H}} \right\|_2^2]$ quantify the expected quadratic differences between the original and the augmented hodge embeddings. \footnote{To see that these indeed depend on $\mathbf{p}$ use the equality $\mathrm{Tr}\left(\bbX \bbX^{\top}\right)=\|\bbX\|_2^2$ and recall $\mathbb{E} \left[\mathbf{x} \circ \mathbf{e} \right] = \mathbf{x} \circ \mathbf{p}$. Then $\mathbb{E} [\left\|\tilde{\mathbf{x}}_{\mathrm{G}} -\tilde{\mathbf{x}}^{\prime}_{\text{G}} \right\|_2^2 ] =  
\left\|\tilde{\mathbf{x}}_{\text{G}}\right\|_2^2  -  \mathrm{Tr}\left( \mathbf{U}_{\mathrm{G}}\mathbf{x}(\mathbf{x} \circ \mathbf{p})^{\top} \mathbf{U}_{\mathrm{G}}^{\top} \right)  - \mathrm{Tr}\left( \mathbf{U}_{\mathrm{G}}(\mathbf{x} \circ \mathbf{p})\mathbf{x}^{\top} \mathbf{U}_{\mathrm{G}}^{\top} \right)  + \mathrm{Tr}\left( \mathbf{U}_{\mathrm{G}}\mathbf{P}\mathbf{U}_{\mathrm{G}}^{\top} \right) $, where $\mathbf{P}$ is a matrix with entries $\mathbf{P}_{i,j}=\mathbf{x}_{i} \mathbf{x}_{j} \mathbf{p}_{i}\mathbf{p}_{j}$ for $i\neq j$ and $\mathbf{P}_{i,j} = (\mathbf{x}_{i})^{2} \mathbf{p}_{i}$ for $i=j$.}
 We use these distances to optimize the probabilities based on prior knowledge such that one or more of the augmented embeddings are similar while the remaining ones differ. For example, when the curl and harmonic embeddings are important, such as in the trajectory prediction task that we will touch in the experiments (Sec.~\ref{sec:num_re}), we could design $\bbp$ by solving:
\begin{subequations}\label{eq.positiveOpt}
\begin{alignat}{2}
&\!\min_{\mathbf{p}}        &\qquad& - \mathcal{L}_{\mathrm{G}}(\mathbf{p}) + \mathcal{L}_{\mathrm{C}}(\mathbf{p}) + \mathcal{L}_{\mathrm{H}}(\mathbf{p}) \label{eq:optProb}\\
&\text{subject to} &      & \mathbf{p} \in \mathcal{G}_{\mathbf{p}}:=\left\{\mathbf{p} \mid \mathbf{p} \in [0,1]^{N_{1}} ,\|\mathbf{p}\|_1 \leq \epsilon_{\mathbf{p}} \right\} ,\label{eq:constraint1}
\end{alignat}
\end{subequations}
where set $\mathcal{G}_{\mathbf{p}}$ puts a maximum budget $\epsilon_{\mathbf{p}}$ on the allocated drop probabilities in a sparse manner (i.e., the $\ell_1-$norm $\|\cdot\|_1$ imposes sparse probabilities on a few flows). The budget  $\epsilon_{\mathbf{p}}$ is tuned as a hyperparameter. \footnote{In \eqref{eq:optProb}, we could also weight the contributions of the different components (e.g., $\alpha_{\rm C}\mathcal{L}_{\mathrm{C}}(\mathbf{p}) + \alpha_{\rm H}\mathcal{L}_{\mathrm{H}}(\mathbf{p})$ with scalars $\alpha_{\rm C}, \alpha_{\rm H} > 0$ when the signals have some contribution in each of them.} By solving \eqref{eq.positiveOpt}, we find the dropout probabilities $\mathbf{p}$ that generate examples with similar curl and harmonic components to the original data point but with a different gradient component. We solve this optimization problem with projected gradient descent, projecting $\mathbf{p}$ onto the constraint set $\ccalG_{\bbp}$ after every step.


\vspace{-.15cm}
\subsection{Hodge-Aware Debiasing}

Problem~\eqref{eq.positiveOpt} influences the embedding space by acting on the augmentation functions $\ccalT_{1\backslash 2}(\cdot)$ to generate better positive examples. To further improve the organization of the embedding space, we shall also act on the negative samples. This is known as a \emph{debiasing} technique and consists of reweighting the negative samples in the InfoNCE loss \cite{chuang2020debiased,Sun2022topkdebiased,LiuGCNDebiased}. I.e., by optimizing w.r.t. the loss:
\begin{equation}\label{eq.overSSLweighLoss}
\mathcal{L}_{\text {weighted }}=-\sum_{\left(\mathbf{x}_{i}^{\prime}, \mathbf{x}_{j}^{\prime}\right) \in \ccalP} \log \left(\frac{e^{ {\rm sim}\left(\mathbf{z}_i^{\prime}, \mathbf{z}_j^{\prime}\right) / \tau }}{\sum_{m=1}^M w(\mathbf{x}_i,\mathbf{x}_m) \ e^{{\rm sim}\left(\mathbf{z}^{\prime}_i, \mathbf{z}_m\right) / \tau }}\right)
\end{equation}
where $w(\mathbf{x}_i,\mathbf{x}_m)$ is the weighting term between the anchor $\bbx_i$ and the negative example $\bbx_m$. For Hodge-aware learning, this weight should reflect the spectral properties of the data; thus, we would like to push spectrally different samples further away from the anchor. 

We first consider a weighted embedding similarity between two data points:
\begin{align}\label{eq.similarity}
\begin{split}
    \mathcal{S}(\tilde{\mathbf{x}}_i, \tilde{\mathbf{x}}_m)= \gamma_{\rm H} \ \mathrm{CD}&(\tilde{\mathbf{x}}_{\mathrm{H},i}, \tilde{\mathbf{x}}_{\mathrm{H},m}) + \gamma_{\rm G} \ \mathrm{CD}(\tilde{\mathbf{x}}_{\mathrm{G},i}, \tilde{\mathbf{x}}_{\mathrm{G},m})\\ 
    & + \gamma_{\rm C} \ \mathrm{CD}(\tilde{\mathbf{x}}_{\mathrm{C},i}, \tilde{\mathbf{x}}_{\mathrm{C},m})
\end{split}
\end{align}
$\mathrm{CD}(\tilde{\mathbf{x}}_i,\tilde{\mathbf{x}}_m) = 1 - \frac{\tilde{\mathbf{x}}^\top_i \tilde{\mathbf{x}}_m}{\left\|\tilde{\mathbf{x}}_i\right\|_2\left\|\tilde{\mathbf{x}}_m\right\|_2}$ is the cosine distance and weights $\gamma_{\rm H},\gamma_{\rm G},\gamma_{\rm C} \geq 0$ are picked based on prior knowledge about the task or tuned as hyperparameters. Choosing the cosine distance leads to higher negative values for spectrally more dissimilar examples. Then, we compute the weight as the normalized similarity over the $M$ negative samples:\vskip-.15cm
\begin{equation}\label{eq.simNorm}
w(\mathbf{x}_{i}, \mathbf{x}_{m})= \frac{\mathcal{S}(\tilde{\mathbf{x}}_i, \tilde{\mathbf{x}}_m)}{\sum_{m=1}^{M} \mathcal{S}(\tilde{\mathbf{x}}_i, \tilde{\mathbf{x}}_m)}
\end{equation}
which means that the weights for each datum sum to one and ensures that the loss terms for different data points are comparable.  Summarizing the above, substituting \eqref{eq.similarity} into \eqref{eq.simNorm} and the latter into \eqref{eq.overSSLweighLoss} leads to an overall contrastive loss that pushes spectrally different samples away from the anchor based on this dissimilarity score. This encourages an embedding space that is organized with respect to the Hodge decomposition.\vskip-.15cm


\begin{figure}[t!]
  \centering
  \includegraphics[width=1\linewidth,scale=1]{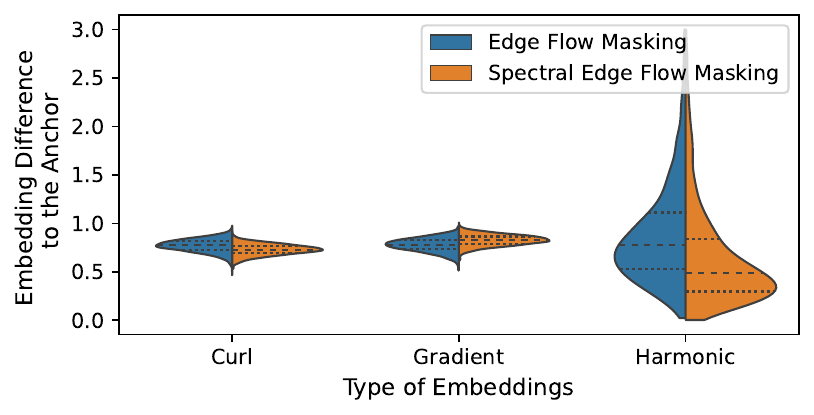}
  \vspace{-7.5mm}
  \caption{Embedding distance for augmentations sampled with uniform probabilities (blue) and with the proposed spectrally optimized ones (orange). In the harmonic embedding, for the distribution associated with the spectral edge drop more probability mass lies over smaller differences and we are thus more likely to generate samples with more similar harmonic components than with the standard method.\vspace{-0.5mm}}\vskip-.15cm
  \label{fig:emb_diff}
\end{figure}

\vspace{-.3cm}
\section{Numerical Results}\vspace{-.15cm}\label{sec:num_re}

We corroborate the proposed approach and compare it with supervised alternatives on two edge flow classification tasks: i) a synthetic task considering trajectories on a map; and ii) a real-data case that contains ocean drifters moving around the Madagascar island. Because of the holes in the SC representations, the harmonic embedding captures important information for solving the task. Due to the limited space, we refer the reader to \cite[\S5]{schaub2020random} for more details.

\smallskip
\noindent\textbf{Setup.} For the trajectory dataset, we generate 200 training, 100 validation, and 100 test data points while there are 160 training and 40 test data points for the ocean drifter. We train the unsupervised simplicial contrastive learner (SCL) on all available unlabled data points and fit a linear support vector machine (SVM) on the obtained embeddings. For the ocean drifters, we use a 10-fold cross-validation on the training set to estimate the penalty parameter for the SVM. We report the average accuracy over 16 data splits. We optimize the network with stochastic gradient descent and grid search the learning rate and weight decay in the interval $[10^{-5}, 1]$ in decimal steps. Furthermore, we select the edge flow drop probabilities $\bbp$ and perturbation budged $\epsilon_{\bbp}$ from $[0.1,0.7]$. All models are trained for 200 epochs with a batch size of 100. For the encoder we follow the setting in \cite{bodnar2021weisfeiler} (which is supervised in there) and use a Tanh-activation and a hidden-layer of size 64. We tune the number of layers and the convolutional orders $L_{1}=L_{2}$ in $[1,2,3]$. We compare the proposed approach with a fully-supervised SCNN and conduct and extensive analysis to understand the role of the different components.

\begin{figure}[t!]
  \centering
  \includegraphics[width=1\linewidth,scale=1]{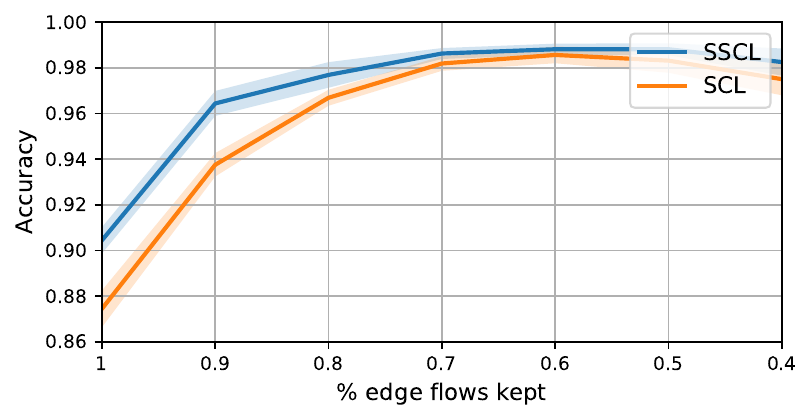}
  \vspace{-7.5mm}
  \caption{Performance comparison between a model trained with the standard loss (SCL) (cf. \eqref{eq.infoNCEloss}) vs. a model trained with a spectrally reweighted loss (SSCL) (cf. \eqref{eq.overSSLweighLoss}) for an edge drop augmentation. The SSCL outperforms the SCL irrespective of the augmentation quality.\vspace{-0.5mm} }\vskip-.15cm
  \label{fig:sscl}
\end{figure}

\smallskip
\noindent\textbf{Results.} Table~\ref{table:test_acc_edge_flows} depicts the overall performance on the downstream tasks. The spectral simplicial contrastive learner (SSCL\textsubscript{Spec}) trained with reweighted negative samples and spectrally optimized probabilities achieves the best downstream accuracy on both datasets. This shows the ability of the proposed approach to effectively encode more relevant Hodge-related information into the embeddings, facilitating the subsequent linear learner. Fig.~\ref{fig:emb_diff} further reinforces this aspect by showing the embedding distance between the anchor and two different augmentation techniques (random edge drop and proposed). The proposed approach generates more similar harmonic embeddings, which is key to the obtained results for the task. In Fig.~\ref{fig:sscl}, we show the proposed approach consistently achieves a superior performance independent of the augmentation quality.

Notably, even for models trained without a reweighted loss, the incorporation of spectrally optimized augmentations (SCL\textsubscript{Spec}) improves the accuracy over uniform probabilities (SCL). This substantiates the importance of the spectral augmentations as a standalone feature. Furthermore, to evaluate the impact of the encoder, we tested a learner that uses only lower Laplacian encoding (SCL$_{\rm low}$), omitting triangle relationships. Compared to its simplicial counterpart SCL under identical conditions, SCL$_{\rm low}$, manifests a noticeable decrease in performance. This demonstrates that the structural advantages of simplicial networks to process flow data transfer to the contrastive learning setting.

 \begin{table}
  \caption{Test Accuracies for the Trajectory and Ocean Drifter datasets. SCL denotes models trained with the standard InfoNCE loss (cf. \eqref{eq.infoNCEloss}), while SSCL models are trained with spectrally reweighted negatives (cf. \eqref{eq.overSSLweighLoss}). The subscript Spec denotes that the augmentation probablities are spectrally optimized.}\vskip-.15cm
  \label{table:test_acc_edge_flows}
  \centering
  \begin{tabular}{lll}
    \toprule
    Model     & Trajectory Task     & Ocean Drifters \\
    \midrule
    SSCL\textsubscript{Spec} (ours)      & $97.9 \pm 0.3$         & $90.3 \pm 1.4$ \\
    SCNN (supervised)                              & $95.2 \pm 	0.5$         & $78.5 \pm 	1.1$ \\
    \midrule
    SSCL      & $96.8 \pm 0.4$         & $89.1 \pm 1.0$ \\
    SCL\textsubscript{Spec}      & $98.2\pm 	0.4$         & $83.1 \pm 1.1$ \\
    SCL     & $96.1 \pm 	0.6$         & $81.6 \pm 	1.6$ \\
    SCL$_{\rm low}$                              & $91.0 \pm 	0.2$         & $77.1 \pm 	1.2$ \\

    \bottomrule
  \end{tabular}\vskip-.15cm
\end{table}

\vspace{-.3cm}
\section{Conclusion}\vskip-.15cm

We show that a contrastive learning framework, when coupled with a simplicial neural network, is effective for generating representations for edge flow data that contain hodge-related information. Related to this, we demonstrated that positive examples with specific spectral properties can be generated by casting the task as an optimization problem on the underlying probabilities of a dropout augmentation. Once these probabilities are optimized, they can be used to generate examples with desired spectral characteristics. We also introduce Hodge-related information into the problem by reweighting the negative examples in the loss based on their spectral difference to the anchor. This pushes spectrally very different examples further apart and results in an embedding space that takes the relevant hodge information into account. Empirical results demonstrate that these optimized embeddings can be used to significantly outperform a fully-supervised model on two edge flow classification tasks. For future work, exploring other types of data augmentation methods and conducting experiments with simplicial complexes of varying dimensions remain as key areas.

\label{sec:concl}

\newpage

\bibliographystyle{IEEEbib}
\bibliography{myIEEEabrv,bib-nonlinear}

\begin{thebibliography}{10}

\bibitem{BATTISTON_20201_pairwise}
F.Battiston, G.Cencetti, I.Iacopini, V.Latora, M.Lucas, A.Patania, J.-G.Young, and G.Petri,
\newblock ``Networks beyond pairwise interactions: Structure and dynamics,''
\newblock {\em Physics Reports}, vol. 874, pp. 1--92, 2020.

\bibitem{robinson2014topological}
M.Robinson,
\newblock {\em Topological signal processing}, vol.~81,
\newblock Springer, 2014.

\bibitem{schaub2021signal}
M.~T.Schaub, Y.Zhu, J.-B.Seby, T.~M.Roddenberry, and S.Segarra,
\newblock ``Signal processing on higher-order networks: Livin'on the edge... and beyond,''
\newblock {\em Signal Processing}, vol. 187, pp. 108149, 2021.

\bibitem{barbarossa2020topological}
S.Barbarossa and S.Sardellitti,
\newblock ``Topological signal processing over simplicial complexes,''
\newblock {\em IEEE Transactions on Signal Processing}, vol. 68, pp. 2992--3007, 2020.

\bibitem{schaub_hodgelets2022}
T.~M.Roddenberry, F.Frantzen, M.~T.Schaub, and S.Segarra,
\newblock ``Hodgelets: Localized spectral representations of flows on simplicial complexes,''
\newblock in {\em ICASSP 2022 - 2022 IEEE International Conference on Acoustics, Speech and Signal Processing (ICASSP)}, 2022, pp. 5922--5926.

\bibitem{isufi_2023_svam}
J.Krishnan, R.Money, B.Beferull-Lozano, and E.Isufi,
\newblock ``Simplicial vector autoregressive model for streaming edge flows,''
\newblock in {\em ICASSP 2023 - 2023 IEEE International Conference on Acoustics, Speech and Signal Processing (ICASSP)}, 2023.

\bibitem{yang_2022_sc_filters}
E.Isufi and M.Yang,
\newblock ``Convolutional filtering in simplicial complexes,''
\newblock in {\em ICASSP 2022 - 2022 IEEE International Conference on Acoustics, Speech and Signal Processing (ICASSP)}, 2022, pp. 5578--5582.

\bibitem{Yang_2021_finite}
M.Yang, E.Isufi, M.~T.Schaub, and G.Leus,
\newblock ``Finite impulse response filters for simplicial complexes,''
\newblock in {\em 2021 29th European Signal Processing Conference ({EUSIPCO})}.

\bibitem{yang_trendfilter_2022}
M.Yang and E.Isufi,
\newblock ``Simplicial trend filtering,''
\newblock in {\em 2022 56th Asilomar Conference onSignals,Systems, and Computers}, 2022.

\bibitem{ebli2020simplicial}
S.Ebli, M.Defferrard, and G.Spreemann,
\newblock ``Simplicial neural networks,''
\newblock in {\em TDA {\&} Beyond}, 2020.

\bibitem{roddenberry2021principled}
T.~M.Roddenberry, N.Glaze, and S.Segarra,
\newblock ``Principled simplicial neural networks for trajectory prediction,''
\newblock in {\em International Conference on Machine Learning}. PMLR, 2021, pp. 9020--9029.

\bibitem{bodnar2021weisfeiler}
C.Bodnar, F.Frasca, Y.~G.Wang, N.Otter, G.Montufar, P.Li{\`o}, and M.~M.Bronstein,
\newblock ``Weisfeiler and lehman go topological: Message passing simplicial networks,''
\newblock in {\em ICLR 2021 Workshop on Geometrical and Topological Representation Learning}, 2021.

\bibitem{simp_con_networks_elvin}
M.Yang, E.Isufi, and G.Leus,
\newblock ``Simplicial convolutional neural networks,''
\newblock in {\em ICASSP 2022 - 2022 IEEE International Conference on Acoustics, Speech and Signal Processing (ICASSP)}, 2022, pp. 8847--8851.

\bibitem{Oord2018}
A.van~den Oord, Y.Li, and O.Vinyals,
\newblock ``Representation learning with contrastive predictive coding,''
\newblock {\em CoRR}, vol. abs/1807.03748, 2018.

\bibitem{SimCLR}
T.Chen, S.Kornblith, M.Norouzi, and G.~E.Hinton,
\newblock ``A simple framework for contrastive learning of visual representations,''
\newblock {\em CoRR}, vol. abs/2002.05709, 2020.

\bibitem{graphcl_2020}
Y.You, T.Chen, Y.Sui, T.Chen, Z.Wang, and Y.Shen,
\newblock ``Graph contrastive learning with augmentations,''
\newblock in {\em Advances in Neural Information Processing Systems}, H.Larochelle, M.Ranzato, R.Hadsell, M.Balcan, and H.Lin, Eds. 2020, vol.~33, pp. 5812--5823, Curran Associates, Inc.

\bibitem{isola_2020_infomin}
Y.Tian, C.Sun, B.Poole, D.Krishnan, C.Schmid, and P.Isola,
\newblock ``What makes for good views for contrastive learning?,''
\newblock in {\em Advances in Neural Information Processing Systems}, H.Larochelle, M.Ranzato, R.Hadsell, M.Balcan, and H.Lin, Eds. 2020, vol.~33, pp. 6827--6839, Curran Associates, Inc.

\bibitem{grace_zhu_2020}
Y.Zhu, Y.Xu, F.Yu, Q.Liu, S.Wu, and L.Wang,
\newblock ``{Deep Graph Contrastive Representation Learning},''
\newblock in {\em ICML Workshop on Graph Representation Learning and Beyond}, 2020.

\bibitem{Zhu_2021}
Y.Zhu, Y.Xu, F.Yu, Q.Liu, S.Wu, and L.Wang,
\newblock ``Graph contrastive learning with adaptive augmentation,''
\newblock in {\em Proceedings of the Web Conference 2021}. apr 2021, {ACM}.

\bibitem{liu2022revisiting}
N.Liu, X.Wang, D.Bo, C.Shi, and J.Pei,
\newblock ``Revisiting graph contrastive learning from the perspective of graph spectrum,''
\newblock in {\em Advances in Neural Information Processing Systems}, A.~H.Oh, A.Agarwal, D.Belgrave, and K.Cho, Eds., 2022.

\bibitem{lin2023spectral}
L.Lin, J.Chen, and H.Wang,
\newblock ``Spectral augmentation for self-supervised learning on graphs,''
\newblock in {\em The Eleventh International Conference on Learning Representations}, 2023.

\bibitem{graph_data_aug_surv_kaize_2022}
K.Ding, Z.Xu, H.Tong, and H.Liu,
\newblock ``Data augmentation for deep graph learning: A survey,''
\newblock {\em SIGKDD Explor. Newsl.}, vol. 24, no. 2, pp. 61–77, dec 2022.

\bibitem{hdd_survey_2013}
H.Bhatia, G.Norgard, V.Pascucci, and P.-T.Bremer,
\newblock ``The helmholtz-hodge decomposition—a survey,''
\newblock {\em IEEE Transactions on Visualization and Computer Graphics}, vol. 19, no. 8, pp. 1386--1404, 2013.

\bibitem{chuang2020debiased}
C.-Y.Chuang, J.Robinson, L.Yen-Chen, A.Torralba, and S.Jegelka,
\newblock ``Debiased contrastive learning,''
\newblock 2020.

\bibitem{Sun2022topkdebiased}
Q.Sun, W.Zhang, and X.Lin,
\newblock ``Progressive hard negative masking: From global uniformity to local tolerance,''
\newblock 2022.

\bibitem{LiuGCNDebiased}
Y.Liu, X.Yang, S.Zhou, X.Liu, Z.Wang, K.Liang, W.Tu, L.Li, J.Duan, and C.Chen,
\newblock ``Hard sample aware network for contrastive deep graph clustering,'' 2022.

\bibitem{schaub2020random}
M.~T.Schaub, A.~R.Benson, P.Horn, G.Lippner, and A.Jadbabaie,
\newblock ``Random walks on simplicial complexes and the normalized hodge 1-laplacian,''
\newblock {\em SIAM Review}, vol. 62, no. 2, pp. 353--391, 2020.

\end{thebibliography}

\end{document}